\documentclass[11pt,letterpaper]{article}
\usepackage{emnlp2016}
\usepackage{times}
\usepackage{latexsym}
\usepackage{url}
\usepackage{graphicx}
\usepackage{arydshln}

\usepackage{algorithm}
\usepackage{algpseudocode}

\algnewcommand\algorithmicinput{\textbf{Input}}
\algnewcommand\Input{\item[\algorithmicinput]}

\algnewcommand\algorithmicreturns{\textbf{Returns}}
\algnewcommand\Returns{\item[\algorithmicreturns]}

\usepackage{multirow}

\usepackage[utf8]{inputenc}	
\usepackage{amsmath}
\usepackage{amsfonts}
\usepackage{amssymb}
\usepackage{xspace}
\usepackage{color}

\usepackage{tipa}
\usepackage{CJKutf8}
\usepackage{booktabs}
\usepackage{todonotes}

\emnlpfinalcopy 

\newcommand{\phoible}{{\sc Phoible}\xspace}

\title{Transliteration in Any Language with Surrogate Languages}

\author{Stephen Mayhew \and Christos Christodoulopoulos \and Dan Roth \\
         University of Illinois, Urbana-Champaign\\
	    201 N. Goodwin\\
	    Urbana, Illinois, 61801\\
	    {\tt \{mayhew2, christod, danr\}@illinois.edu}}

\date{}

\begin{document}

\maketitle

\begin{abstract}

We introduce a method for transliteration generation that can produce transliterations in every language. Where previous results are only as multilingual as Wikipedia, we show how to use training data from Wikipedia as surrogate training for any language. Thus, the problem becomes one of ranking Wikipedia languages in order of suitability with respect to a target language. 
We introduce several task-specific methods for ranking languages, and show that our approach is comparable to the oracle ceiling, and even outperforms it in some cases.



\end{abstract}

\section{Introduction}
Transliteration is the task of converting a word or phrase, typically a named entity, from one script into another while maintaining the pronunciation. The problem is largely of interest for downstream tasks such as cross lingual named entity recognition \cite{Darwish2013ACL,kim2012multilingual}. Work in transliteration can be divided into two areas: \textit{discovery}, where a query name is provided along with a list of candidates, and the task is to select the best candidate, and \textit{generation}, where the task is to generate a transliteration for a given test name. We focus on the more challenging task, generation. 

To date, all of the work in generation has required some level of supervision in the target language $T$ in the form of name pairs.  Multilingual-minded approaches typically gather this from Wikipedia \cite{PasternackRo09a,irvine2010transliterating}. However, languages with little or no presence in Wikipedia are still out of reach.


We describe a simple method to use the languages present in Wikipedia to extend transliteration capability to all languages. The intuition is that there are enough languages in Wikipedia such that for any language in the world, we can select a Wikipedia language that is \textit{close enough} to the target language to build a transliteration model. Our method produces results that are consistently comparable or superior to an oracle that selects the best language every time. Further, for some small Wikipedia languages, we can produce results better than training on the target language.







\begin{figure}
\centering
\includegraphics[scale=0.6]{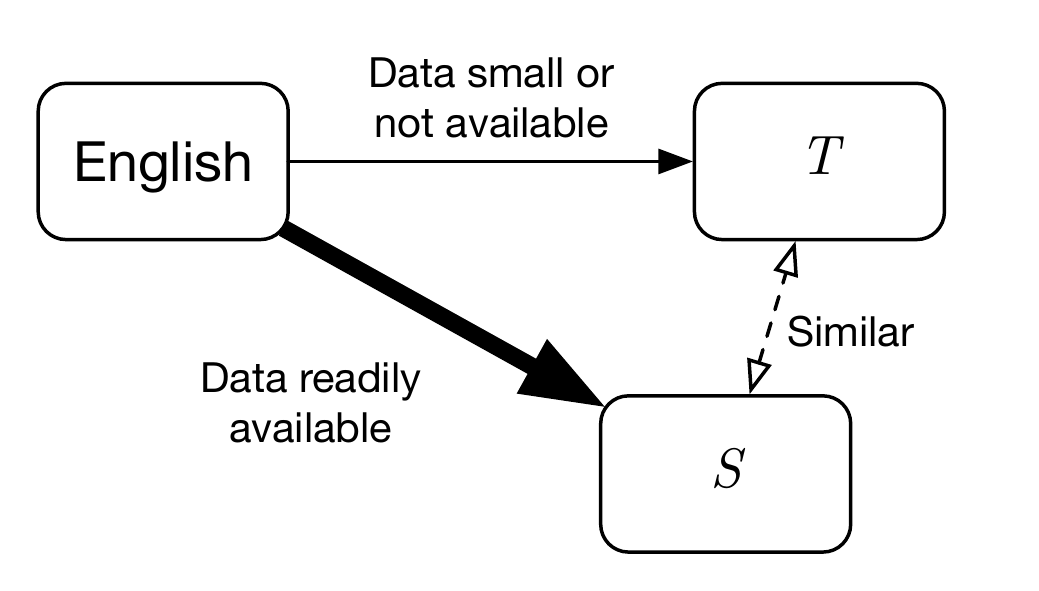}
\caption{Transliteration into target language $T$. The challenge in this setup is in selecting an appropriate surrogate language $S$.}
\end{figure}

\section{Method}

Central to our approach is the fact that we have name pair data from 281 languages in Wikipedia. Of these, many have a trivial number of name pairs ($<$ 20), so we keep only those 131 languages with 200 name pairs or more. We refer to this set as $\mathcal{W}$. 

We used uroman\footnote{\url{http://www.isi.edu/natural-language/software/romanizer/uroman-v0.4.tar.gz}} to romanize all of our data. This means that the text in each language is represented using Latin characters, and that we can train on any language, and use that model to test on any other language, unconstrained by script. Note that romanization is a deterministic process that is reversible with little loss.\footnote{uroman does not currently support this.}

Our goal is to generate transliterations from English into some (potentially low-resource) target language $T$. We use English because of the availability of resources, although in principle our method works for any source language with enough resources. One application is to use English to search for named entities in a large corpus in language $T$. We also note that transliteration is symmetric. Given name pairs, one can learn a model in either direction.


Now, given $T$, there are two situations: (1)~ $T$ is in $\mathcal{W}$, or (2)~ $T$ has little or no data in Wikipedia. In the first case, if $T$ has a sufficient number of training pairs, then we can simply use that data to train. However, if it has a small number of training pairs, or if it is not in Wikipedia, we need to resort to more interesting methods. 

The method we propose is shown in Algorithm \ref{alg:tl}.  The interesting part is in selecting a \textit{surrogate} language $S$ from $\mathcal{W}$. We call it a \textit{surrogate} because it acts exactly as though it were $T$. This selection is built around the notion of language similarity, as seen in the $sim()$ function.




For all of our experiments, we use the transliteration model described in \newcite{PasternackRo09a}. This is a probabilistic model that allows the mapping of variable length segments between languages. In all following experiments we used 5 EM iterations, with a maximum segment length of 15 for source and target words, and \texttt{SegmentFactor} of 0.5.

\begin{algorithm}[t]
\caption{Our method}
\label{alg:tl}
\begin{algorithmic}[1]
\Input 
\Statex $N$: set of names in English
\Statex $T$: target language
\Returns
\Statex $N'$: names transliterated into $T$
\For{$\forall L \in \mathcal{W}$} \Comment Surrogate selection
\State Get $sim(L, T)$
\EndFor
\State Order $\mathcal{W}$ according to scores
\State $S =$ top language in $\mathcal{W}$
\State Train model using $(English, S)$
\State Use model to transliterate $N$ into $N'$
\end{algorithmic}
\end{algorithm}


\section{Transliteration-Driven Language Similarity}
\label{sec:sim}

We start by presenting the supervised oracle ranking, and then describe our proposed similarity metrics. 

\subsection{Oracle Ranking}
In order to evaluate our proposed similarity metrics, we need to have an oracle ranking. We gather this oracle ranking with respect to language $T$ as follows. For each language $L$ in Wikipedia (including $T$, if it is there), we learn a model, and test this model on name pairs in language $T$. Now we have a score for each $L$, which we use to induce a ranking over $\mathcal{W}$. This is the oracle ranking.

We would expect that $T$ is ranked at the top, and languages much different from $T$ are near the bottom. For example,  a model trained on Russian should perform well on Ukrainian data, but a model trained on Korean should not. Note that in order to get the ranking, you need name pairs from $T$.





\subsection{Phonetic Similarity}

Our phonetic similarity metric is based on phonetic inventories, under the intuition that languages with similar inventories will have similar transliterations. This addresses the second challenge from \newcite{KarimiScTu11}, that of missing sounds. 

We use \phoible \cite{phoible}, a database of phonetic inventories of over 1600 languages, each composed of a common set of symbols from the International Phonetic Alphabet. We compute similarity between two languages $X$ and $Y$ by measuring the $F_1$ between phoneme inventory sets $L_X$ and $L_Y$. We refer to this as $sim_{phon}(X,Y)$.


\subsection{Script Distribution Similarity}
We calculate the script similarity of languages $X$ and $Y$ as the cosine similarity of character histogram vectors $H_X$ and $H_Y$. We romanize and lowercase all text first, so these vectors usually have length about 26 (some languages include spurious characters, and some do not use all 26 characters). Any small body of target text will suffice to estimate $H_X$ and $H_Y$. We refer to this as $sim_{script}(X,Y)$.


\subsection{Genealogical Similarity}
WALS \cite{wals} provides some basic features of nearly every language, including the language family and genus \cite{wals-gll}, where family is a broader distinction than genus. We calculate the genealogical similarity of languages $X$ and $Y$ as follows: $1$, if genus and family match, $0.5$ if just family matches, and 0 otherwise. We call this $sim_{gen}(X,Y)$

\subsection{Learned Similarity}

We combined the component similarities by using them as features in $\text{SVM}^{\textit{rank}}$ \cite{joachims2006training}, and using the oracle rankings as supervision. In addition to the component similarities, we also used all 6 distance features from URIEL \cite{LittelMoLe16}, a collection of language resources. These distances are calculated from features in WALS and {\sc Phoible}, and are distributed with the URIEL package. The distance features are: genetic, geographic, inventory, phonological, syntactic, and featural. When predicting rankings on any target language, we trained on all the other experimental languages (see \S~\ref{sec:exp}). We refer to this as $sim_{learned}(X,Y)$.




\section{Experiments and Results}
\label{sec:exp}

We chose 9 languages to act as low-resource target languages (although some of them are actually high-resource): Bengali, Chuvash, Armenian (hye), Kannada, Mazandarani, Newar, Thai, Uzbek, and Mingrelian (xmf). We included a diverse set of scripts, and we tended towards languages with a small number of training pairs. 


To test our hypotheses, we did three experiments. The first, in 
\S~\ref{sec:exp1}, validates the similarity metric. The second, in \S~\ref{sec:exp2}, shows that using surrogate languages can be effective. The third, in \S~\ref{sec:exp3}, shows that combining multiple surrogate languages can be even more effective.










\begin{table*}
\centering
\begin{tabular}{lllllllllll}
\toprule
Method $\backslash$ Lang & ben & chv & hye & kan & mzn & new & tha & uzb & xmf & Avg. \\
\midrule
Random & 26.13 & 58.88 & 38.97 & 41.32 & 9.65 & 24.24 & 27.72 & 48.71 & 66.82 & 38.05 \\
$sim_{phon}$ & 86.86 & 75.50 & 86.18 & \textbf{98.81} & 62.81 & 31.67 & 78.87 & 47.10 & 61.04 & 69.87 \\
$sim_{script}$ & 82.60 & \textbf{97.11} & 92.41 & 94.05 & \textbf{99.93} & \textbf{76.70} & \textbf{97.48} & \textbf{94.46} & \textbf{97.29} & \textbf{92.45} \\
$sim_{gen}$ & \textbf{90.09} & 82.58 & \textbf{97.68} & 87.2 & 61.11 & 23.80 & 95.77 & 91.43 & 84.82 & 79.39 \\
$sim_{learned}$ & 76.44 & 92.45 & 97.01 & 89.50 & 99.65 & 70.19 & 96.39 & 91.91 & 77.61 & 87.91 \\
\bottomrule
\end{tabular}
\caption{Ranking comparison results, showing scores for different ranking methods. Each ranking method is evaluated against the oracle ranking. All scores NDCG with $k=5$. The top two rows are baselines. Script similarity is clearly the best ranking metric.}
\label{tab:sim}
\end{table*}

\subsection{Similarity Metric Validation}
\label{sec:exp1}

In the first experiment, we compare our similarity rankings against the oracle ranking, using normalized discounted cumulative gain (NDCG), a metric commonly used in Information Retrieval \cite{jarvelin2002cumulated}. In this setting, an IR system ranks a set of documents, each of which has a pre-defined relevance score. This captures the notion that highly relevant documents should be placed at the top of a retrieved list. The normalized version (NDCG) divides each DCG score by the maximum possible score. In our situation, documents correspond to languages, and the relevance score of each language $S$ is the score achieved when we train on $S$ and test on $T$ (i.e. scores from the oracle ranking). We set $k=5$, which means that we expect relevant languages to appear in the first 5 results. 




Table \ref{tab:sim} shows results. The first row shows scores from a random permutation of the relevance scores. Each row following that shows results from the similarity metrics described in \S~\ref{sec:sim}.

The clear winner is the script distance, in the third row. This is a surprising result because it is the least knowledge-intensive metric. We had expected that the phonetic similarity would be more successful. Although it performs well for Kannada, it produces dismal scores for others.

$sim_{learned}$ performs well on average, but never attains the best score for a language. We found that script similarity is by far the strongest feature, with a weight of about 30 times that of other weights.

\begin{table*}
\centering
\begin{tabular}{lllllllllll}
\toprule
Method $\backslash$ Lang & ben & chv & hye & kan & mzn & new & tha & uzb & xmf & Avg. \\
\midrule
Trained on $T$ & 23.90 & 57.02 & 58.54 & 24.75 & 44.67 & 7.63 & 29.60 & 44.73 & 51.59 & 38.05 \\
Oracle top & 10.47 & 61.51 & 23.17 & 26.22 & 41.84 & 23.16 & 8.30 & 43.58 & 56.20 & 32.72 \\
\midrule
Predicted top & 6.22 & 58.33 & 19.61 & 26.22 & 41.84 & 15.88 & 8.30 & 40.05 & 56.20 & 30.29 \\
Predicted top 5 & 8.40 & 61.34 & 22.43 & 21.70 & 35.10 & 24.81* & 7.99 & 44.37* & 45.68 & 30.20 \\
\bottomrule
\end{tabular}
\caption{Scores when using top $k$ candidate languages for $T$. For the top $k$, we always exclude $T$ from the list. All scores are MRR. *indicates the score is better than the oracle}
\label{tab:mrr}
\end{table*}

\subsection{Effectiveness of Surrogate Languages}
\label{sec:exp2}
In the second experiment, we select the top language (as predicted by script similarity), and show that it can produce transliterations comparable in quality to the best language. 

The results are shown in Table \ref{tab:mrr}, where every value is an MRR score. The `Oracle top' row shows scores from training on the top oracle language, and the `Predicted top' shows the score from the top predicted language. The oracle score is, by definition, always greater than or equal to the predicted ranking. Note that we removed $T$ from all rankings in order to simulate a low-resource setting.

The top predicted language is the same as the oracle in 4 out of 9 cases. In each other case, the score of the top predicted language is not far from the best. 

The exception is Newar (new), which is 8 points below the best. We hypothesize that this is because the script similarity score doesn't take into account the size of the training data. The top predicted language, Nepali, has a small number of training pairs.  

\subsection{Combining Surrogate Languages}
\label{sec:exp3}
The third experiment combines the top $k$ predicted languages into a single prediction result via weighted voting. At test time, each predicted language produces an $n$-best list with scores for each word in the test set. We combine $n$-best lists by using the score to vote. The results are seen in Table \ref{tab:mrr}, in the row titled `Predicted top $k$', where $k=5$. 

In all cases, if the predicted top is not the best choice (for example, as in Uzbek), then the combination score helps. In two cases, Newar and Uzbek, the top 5 combination gives a score that is better than the oracle (indicated by a *). 



\section{Related Work}
\label{sec:wave}

Transliteration is typically studied in the context of a single language or a group of languages. To name a few, there is work on Arabic \cite{sherif2007substring} and Japanese \cite{KnightGr98}. For an excellent survey on the state of the art in transliteration, see \newcite{KarimiScTu11}. 

\newcite{PasternackRo09a} and \newcite{IrvineWeCa12} both do transliteration using data harvested from Wikipedia, but cannot address those languages with little or no presence in Wikipedia. 

There is work on transliteration using only monolingual phonetic mappings \cite{jagarlamudi2012regularized,yoon2007multilingual}. While these mappings are less expensive than name pairs, they still require expert knowledge to create. 

We contrast our similarity metric with a \textit{transliterability} measure called Weighted AVerage Entropy (WAVE) \cite{kumaran2010compositional}. WAVE measures the frequency of alphabet ngrams weighted by the entropy of the ngram mapping. However, to estimate WAVE, one needs to have a set of name pairs in $T$.

\newcite{Rosa2015KLcpos3A} introduce a language similarity metric for selecting a source treebank in cross lingual parsing. The metric is based on KL-divergence of POS trigrams, and is analogous to WAVE in that annotations in $T$ are required. 

In the multilingual tradition, this paper is closely related to direct transfer techniques \cite{tackstrom2012cross,mcdonald2011multi}.

\section{Discussion and Future Work}

We have shown a way to extend transliteration capabilities into all languages using Wikipedia as a data source. We showed a method for using task-driven language similarity metrics to induce a ranking over languages that is very close to an oracle ranking. Finally, our combination algorithm is capable of producing scores that outperform the oracle scores.

In the future, we will explore more sophisticated techniques for combining languages, with the hope of significantly beating the oracle every time. 





\bibliography{notes,ccg,cited}
\bibliographystyle{emnlp2016}

\end{document}